\pdfoutput=1
\documentclass[11pt]{article}

\usepackage{ACL2023}

\usepackage{times}
\usepackage{latexsym}

\usepackage[T1]{fontenc}
\usepackage[utf8]{inputenc}

\usepackage{microtype}

\usepackage{inconsolata}

\usepackage{booktabs}
\usepackage{multirow}
\usepackage{graphicx} % for table scaling
\usepackage{adjustbox} % for better table fitting
\usepackage{pgfplots}
\usepackage{tikz}
\usepackage{subcaption}

\title{Evaluating the Impact of Synthetic Data on Object Detection Tasks in Autonomous Driving}

\author{
  Enes Özeren \and Arka Bhowmick \\
  BIT Technology Solutions GmbH \\
  \texttt{enes.oezeren@bit-ts.de} \\
  \texttt{arka.bhowmick@bit-ts.de} \\
}

\pgfplotsset{compat=1.18}
\begin{document}

\maketitle
\begin{abstract}
The increasing applications of autonomous driving systems necessitates large-scale, high-quality datasets to ensure robust performance across diverse scenarios. Synthetic data has emerged as a viable solution to augment real-world datasets due to its cost-effectiveness, availability of precise ground-truth labels, and the ability to model specific edge cases. However, synthetic data may introduce distributional differences and biases that could impact model performance in real-world settings. To evaluate the utility and limitations of synthetic data, we conducted controlled experiments using multiple real-world datasets and a synthetic dataset generated by BIT Technology Solutions GmbH. Our study spans two sensor modalities, camera and LiDAR, and investigates both 2D and 3D object detection tasks. We compare models trained on real, synthetic, and mixed datasets, analyzing their robustness and generalization capabilities. Our findings demonstrate that the use of a combination of real and synthetic data improves the robustness and generalization of object detection models, underscoring the potential of synthetic data in advancing autonomous driving technologies.
\end{abstract}

\section{Introduction}

Autonomous driving systems are complex agents that make decisions to operate safely and effectively in real-world environments. At the core of these systems are machine learning algorithms, which process sensor inputs and guide decision-making. Broadly, there are two primary approaches to leveraging machine learning in these systems: modular perception planning and end-to-end learning \citep{grigorescu2020survey}. In the end-to-end approach, a single machine learning algorithm takes input directly from sensors and determines actions. In contrast, the modular perception planning approach breaks the problem into subtasks handled by dedicated machine learning models and classical algorithms. One critical subtask in modular perception planning is perception, which involves interpreting sensor data to understand the surrounding environment. Two major tasks within perception are 2D and 3D object detection.

2D and 3D object detection aim to localize and identify objects in the environment. In 2D object detection, the task is to predict bounding boxes around objects in images captured by cameras, whereas 3D object detection involves predicting bounding boxes in three-dimensional space, often using point cloud data generated by LiDAR sensors. The primary input for 2D detection is image data, while 3D detection typically relies on point clouds due to their detailed spatial information.

To train object detection models effectively, machine learning algorithms require large datasets with accurate labels. For 2D object detection, datasets consist of images annotated with 2D bounding boxes, while 3D object detection datasets include point clouds annotated with 3D bounding boxes. However, collecting these datasets from real-world environments has significant challenges:

1. Time-Consuming Data Collection: Gathering data using vehicles equipped with sensors requires extensive time and effort.

2. High Costs: Sensors such as cameras and LiDAR, as well as the labor required for data collection, are expensive.

3. Labor-Intensive Labeling: Annotating collected data with high-quality labels is both time-consuming and costly.

Due to these challenges, acquiring sufficient real-world datasets can be difficult. A promising solution to this problem is the use of synthetic data, which can be generated in simulated environments. Synthetic data offers several advantages, such as reduced cost, availability of perfectly accurate labels, and the ability to create diverse scenarios, including rare edge cases. However, the quality and biases inherent in synthetic datasets are critical considerations, as poorly generated synthetic data can lead to suboptimal model performance.

In this study, we conduct controlled experiments to assess the utility of synthetic data generated by BIT Technology Solutions GmbH (BIT-TS). The synthetic dataset used in our experiments includes images, point clouds, and corresponding ground-truth labels created using BIT-TS's simulation and data generation systems. We train 2D and 3D object detection models using various real-world datasets, the synthetic dataset, and combinations of both, and we compare the results to evaluate robustness and generalization across different data modalities and detection tasks.

\section{Related Work}

The increasing demand for high-quality training data has led to a growing interest in synthetic datasets, which offer cost-effective and efficient alternatives to real-world data. The evaluation of the quality and utility of synthetic data has thus become a critical research focus.

A comprehensive survey by \citet{song2023synthetic} provides an extensive overview of various synthetic datasets, analyzing their characteristics such as dataset volume, diversity of scene conditions, sensor types, and label formats. The study also explores the performance of 2D object detection models trained on synthetic data and tested on real-world datasets. However, it is limited to 2D detection tasks and does not investigate the impact of combining synthetic and real data. Additionally, the study lacks evaluation experiments for 3D object detection or point cloud-based datasets, which are crucial for many autonomous driving applications.

\citet{talwar2020evaluating} conducted experiments using YOLOv3 \citep{redmon2018yolov3}, comparing the performance of models trained on synthetic datasets versus real datasets when tested on real-world data. While this research offers valuable insights, it is restricted to 2D object detection on images and does not include 3D object detection tasks or point cloud data. Moreover, the detection tasks were limited to vehicles, excluding other critical object classes such as pedestrians. Additionally, the study did not explore the effects of mixing synthetic and real datasets during training, which could potentially improve model generalization. 

Our study addresses some of these gaps by conducting experiments across both 2D and 3D object detection tasks, utilizing synthetic and real datasets, as well as their combinations. By incorporating vehicle and pedestrian object classes and using both image and point cloud data, our work provides a more comprehensive evaluation of the utility and limitations of synthetic datasets in autonomous driving scenarios.

\section{Methodology}

\subsection{Datasets}

\begin{figure}[htbp] % Position: here, top, bottom, or page
    \centering
    \includegraphics[width=0.47\textwidth]{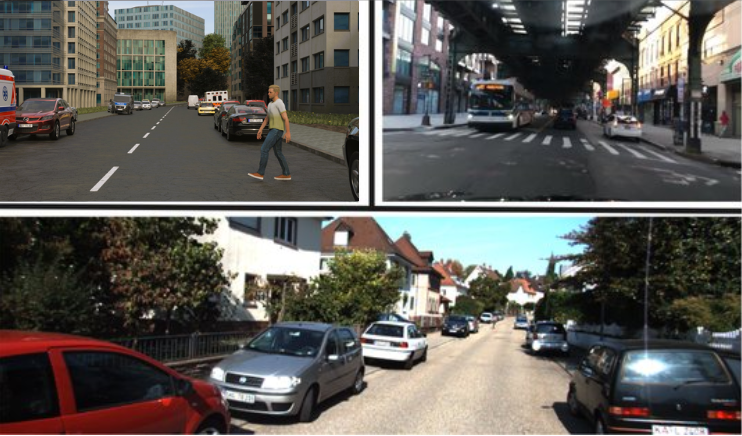}
    \caption{Image examples from three datasets: BIT-TS synthetic data (top left), BDD100K (top right), and KITTI (bottom).}
    \label{fig:image_samples}
\end{figure}

We evaluate the performance of 2D and 3D object detection models trained on real, synthetic, and mixed datasets. For the 2D object detection task, we utilize image data captured by camera sensors from the KITTI \citep{geiger2013vision} and BDD100K \citep{yu2020bdd100k} datasets as real data, and the BIT-TS dataset as synthetic data (see Figure~\ref{fig:image_samples}).

\begin{figure}[htbp] % Position: here, top, bottom, or page
    \centering
    \includegraphics[width=0.47\textwidth]{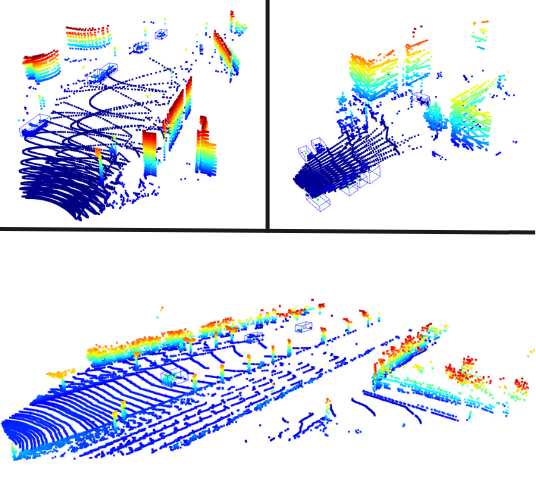}
    \caption{Point Cloud examples from three datasets: BIT-TS synthetic data (top left), A2D2 (top right), and KITTI (bottom).}
    \label{fig:pointcloud_samples}
\end{figure}

For the 3D object detection task, we utilize point cloud data from LiDAR sensors, using KITTI and A2D2 \citep{geyer2020a2d2} as real datasets, and BIT-TS as the synthetic dataset. Since the BDD100K dataset does not include LiDAR point cloud data, it is replaced by A2D2 for 3D object detection experiments (see Figure~\ref{fig:pointcloud_samples}).

The volume of frames utilized for each dataset is detailed in Table~\ref{datasets-and-volumes}. 

\begin{table}[h!]
   \small
   \centering
   \begin{tabular}{lccc}
   \toprule\toprule
   \textbf{Dataset Type} & \textbf{Dataset} & \multicolumn{2}{c}{\textbf{Utilized Frame Count}} \\ 
   \cmidrule(lr){3-4}
                         &                  & \textbf{Image} & \textbf{Point Cloud} \\ 
   \midrule
   \multirow{3}{*}{Real} & KITTI            & 6000           & 4000                 \\
                         & BDD100K          & 6000           & -                    \\
                         & A2D2             & -              & 4000                 \\ 
   \midrule                                     
   Synthetic             & BIT-TS           & 6000           & 4000                 \\                   
   \bottomrule
   \end{tabular}
   \caption{\label{datasets-and-volumes}
The real and synthetic datasets utilized for training the detection models. These volumes include the train and validation sets but excludes the test sets. Images are from camera and point clouds are from LiDAR sensor. All numbers represent the number of frames.
}
\end{table}

The KITTI dataset was collected in Karlsruhe, Germany, using a vehicle equipped with sensors. The data was primarily captured during daytime under clear weather conditions, providing approximately 7,000 labeled image and point cloud frames \citep{geiger2013vision}. In contrast, the BDD100K dataset was collected from various urban streets across the United States, featuring a wider range of weather and environmental conditions \citep{yu2020bdd100k}. Although the BDD100K dataset is significantly larger than KITTI and BIT-TS, we utilized only a random subsample in our experiments to ensure consistency across datasets.

Similarly, the A2D2 dataset, which was collected in multiple cities across Germany, offers a larger data volume than what was used in our experiments. Lastly, the BIT-TS dataset is a synthetic dataset created by BIT Technology Solutions GmbH, based in Munich, Germany. BIT-TS contains highly realistic driving scenarios with pre-generated ground truth labels, making it a valuable synthetic dataset for benchmarking.

To ensure fairness and consistency in our experiments, we limited the number of utilized frames to match the smallest dataset available: KITTI for image data and BIT-TS for point cloud data, as detailed in Table~\ref{datasets-and-volumes}.

\subsection{Object Detection Models}

For the 2D and 3D object detection tasks, we utilize the YOLOv7 \citep{wang2023yolov7} and SECOND \citep{yan2018second} models, respectively. These models were selected for two primary reasons: first, they are well-established algorithms with a proven track record of success in object detection; and second, their open-source repositories facilitate ease of implementation.

YOLOv7 is an advanced real-time object detection algorithm designed for high-speed and accurate detection \citep{wang2023yolov7}. It builds on the YOLO family of models, introducing several optimizations to improve performance and efficiency. The YOLOv7 family consists of different models, each with varying parameter sizes and architectural differences. For our experiments, we selected the mid-sized variant, YOLOv7-w6, which supports input resolutions up to 1280x1280 pixels. We use the original Github code repository for training the YOLOv7 model \footnote{https://github.com/WongKinYiu/yolov7}.

SECOND (Sparse Convolutional Detection) is a 3D object detection algorithm that operates on LiDAR point clouds \citep{yan2018second}. It leverages sparse convolutional neural networks (Sparse CNNs) for efficient processing of high-dimensional point cloud data. We use the OpenPCDet Github code repository for training the SECOND model \citep{openpcdet2020} \footnote{https://github.com/open-mmlab/OpenPCDet}.

\subsection{Metrics}

For both of the 2D and 3D object detection tasks, we use the well-known and expressive metric mean average precision at 50 (mAP@50). The mAP metric evaluates the performance of object detection models by computing the average precision (AP) across all classes, with the intersection-over-union (IoU) threshold set to 0.5.

The formula for mean average precision is given by:

\[
mAP = \frac{1}{N} \sum_{i=1}^{N} AP_i
\]

where \(N\) is the number of classes, and \(AP_i\) represents the average precision for class \(i\). The average precision is calculated by first computing the precision-recall curve and then integrating the curve to obtain the area under the curve (AUC).

For all of our experiments, we use only \textbf{Pedestrian} and \textbf{Car} objects, which are the two common objects present in all datasets. These objects are of particular importance as they are the main focus in the context of safety, and thus, they are critical for evaluating the effectiveness of object detection models in real-world scenarios. Therefore, we only give the labels for these 2 categories to the models during training and present the mAP@50 values for these 2 categories.

\section{Experiments}

\subsection{2D Object Detection}

For the 2D object detection task, we trained 7 models, each using a different dataset while keeping the hyperparameters consistent across all models. Table~\ref{2d-detection-datasets} presents the volumes of the training datasets used for each of the 7 models.

\begin{table*}[h]
\centering
\begin{adjustbox}{valign=c} % This ensures table fits better when necessary
\resizebox{\textwidth}{!}{%
\begin{tabular}{|l|l|c|c|c|c|}
\hline
\textbf{Model Name} & \textbf{Dataset} & \textbf{BIT Frames} & \textbf{KITTI Frames} & \textbf{BDD100K Frames} & \textbf{Total Frames} \\
\hline
2D Model 1 & BIT-TS & 6K & 0 & 0 & 6K \\
2D Model 2 & KITTI & 0 & 6K & 0 & 6K \\
2D Model 3 & BIT-TS \& KITTI MIX & 2K & 6K & 0 & 8K \\
2D Model 3+ & BIT-TS \& KITTI MIX+ & 4K & 6K & 0 & 10K \\
2D Model 3++ & BIT-TS \& KITTI MIX++ & 6K & 6K & 0 & 12K \\
2D Model 4 & BDD100K & 0 & 0 & 6K & 6K \\
2D Model 4++ & BIT-TS \& BDD100K MIX++ & 6K & 0 & 6K & 12K \\
\hline
\end{tabular}
}
\end{adjustbox}
\caption{2D Object Detection - YoloV7: Training datasets for the 7 models (Note: "K" denotes thousands).}
\label{2d-detection-datasets}
\end{table*}

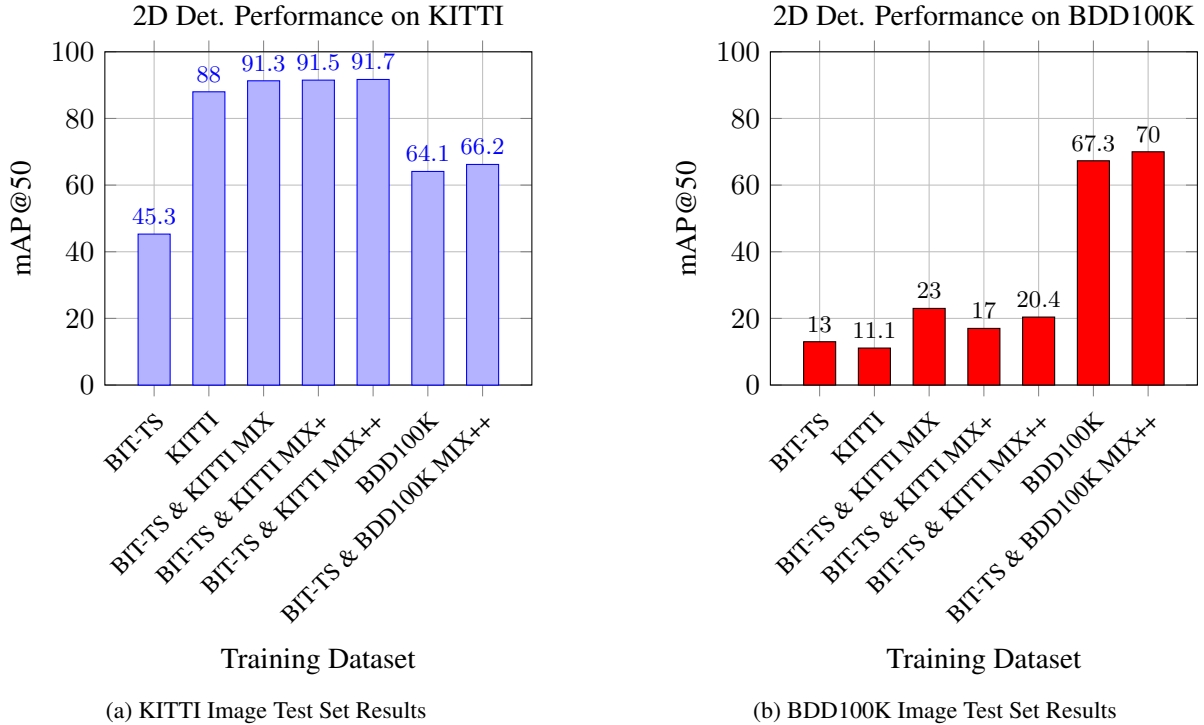
\begin{figure*}[h!]
    \centering
    \begin{subfigure}[b]{0.45\textwidth}
        \centering
        \begin{tikzpicture}
            \begin{axis}[
                width=\textwidth,
                height=6cm,
                ybar=2pt,
                xtick=data,
                symbolic x coords={BIT-TS, KITTI, BIT-TS \& KITTI MIX, BIT-TS \& KITTI MIX+, BIT-TS \& KITTI MIX++, BDD100K, BIT-TS \& BDD100K MIX++},
                ylabel={mAP@50},
                xlabel={Training Dataset},
                bar width=12pt,
                enlarge x limits=0.15,
                grid=major,
                yticklabel style={/pgf/number format/.cd, fixed, precision=0},
                nodes near coords,
                nodes near coords style={font=\footnotesize},
                x tick label style={rotate=45,anchor=north east,font=\footnotesize},
                title={2D Det. Performance on KITTI},
                ymin=0,
                ymax=100
            ]
            \addplot coordinates {(BIT-TS,45.3) (KITTI,88.0) (BIT-TS \& KITTI MIX,91.3) (BIT-TS \& KITTI MIX+,91.5) (BIT-TS \& KITTI MIX++,91.7) (BDD100K,64.1) (BIT-TS \& BDD100K MIX++,66.2)};
            \end{axis}
        \end{tikzpicture}
        \caption{KITTI Image Test Set Results}
        \label{fig:2d-kitti}
    \end{subfigure}
    \hfill
    \begin{subfigure}[b]{0.45\textwidth}
        \centering
        \begin{tikzpicture}
            \begin{axis}[
                width=\textwidth,
                height=6cm,
                ybar=2pt,
                xtick=data,
                symbolic x coords={BIT-TS, KITTI, BIT-TS \& KITTI MIX, BIT-TS \& KITTI MIX+, BIT-TS \& KITTI MIX++, BDD100K, BIT-TS \& BDD100K MIX++},
                ylabel={mAP@50},
                xlabel={Training Dataset},
                bar width=12pt,
                enlarge x limits=0.15,
                grid=major,
                yticklabel style={/pgf/number format/.cd, fixed, precision=0},
                nodes near coords,
                nodes near coords style={font=\footnotesize},
                x tick label style={rotate=45,anchor=north east,font=\footnotesize},
                title={2D Det. Performance on BDD100K},
                ymin=0,
                ymax=100
            ]
            \addplot [fill=red] coordinates {(BIT-TS,13.0) (KITTI,11.1) (BIT-TS \& KITTI MIX,23.0) (BIT-TS \& KITTI MIX+,17.0) (BIT-TS \& KITTI MIX++,20.4) (BDD100K,67.3) (BIT-TS \& BDD100K MIX++,70.0)};
            \end{axis}
        \end{tikzpicture}
        \caption{BDD100K Image Test Set Results}
        \label{fig:2d-bdd100k}
    \end{subfigure}
    \caption{Test set 2D object detection mAP@50 values for YOLOv7 - 2D Detection Models}
    \label{fig:2d-detection-results}
\end{figure*}

For training, we initialize the model parameters randomly, avoiding transfer learning to maintain a controlled experimental environment that allows us to assess the impact of the different datasets used. The training process employs the SGD algorithm, with an initial learning rate of 0.007 and a learning rate scaling factor of 0.1. A weight decay of 0.0005 is applied, and the default data augmentations from the YOLOv7 codebase are used \citep{wang2023yolov7}. The batch size is set to 4 per GPU (16 in total), and the model is trained for 50 epochs with 4 Nvidia RTX 6000 GPUs. All 7 models are trained with the same hyperparameters and we observe that 50 epochs are sufficient for all models to converge with respect to the validation loss. We select the model checkpoint corresponding to the best validation loss.

In Figure~\ref{fig:2d-detection-results}, the test results on the KITTI and BDD100K datasets are presented. The test sets were held out from the training and validation sets to ensure an unbiased evaluation.

Subfigure~\ref{fig:2d-kitti} shows the performance of models tested on the KITTI dataset. The model trained exclusively on the BIT-TS synthetic dataset performs poorly. However, combining BIT-TS synthetic data with the KITTI dataset significantly improves performance compared to training only on the KITTI dataset. As the proportion of BIT-TS synthetic data increases in the training set, a gradual improvement in performance is observed, with the best results achieved by the model trained on the BIT-TS \& KITTI MIX++ dataset. Additionally, adding synthetic data to the BDD100K training set improves the model's performance on the KITTI test set, indicating that incorporating synthetic datasets into real datasets can enhance the model's generalization to other cases.

Subfigure~\ref{fig:2d-bdd100k} highlights the results for models tested on the BDD100K dataset. Models that were not trained with any BDD100K data perform poorly, which is expected given the distinct characteristics and biases in the BDD100K dataset compared to others. BDD100K includes challenging weather conditions and diverse scenarios, making it significantly different from datasets like KITTI. However, mixing BIT-TS synthetic data with KITTI data improves the model’s performance on the BDD100K test set. This demonstrates that using synthetic datasets alongside real datasets can enhance the model's generalization to other scenarios, similar to the pattern observed in Subfigure~\ref{fig:2d-kitti}. The best performance is achieved by the model trained on the BIT-TS \& BDD100K MIX++ dataset, which mirrors the trend seen in Subfigure~\ref{fig:2d-kitti}. Interestingly, the model trained exclusively on BIT-TS synthetic data slightly outperforms the one trained on KITTI when evaluated on the BDD100K test set. This result underscores the adaptability of BIT-TS synthetic data to diverse and challenging scenarios.

Overall, the results demonstrate the effectiveness of mixing synthetic and real data in improving model performance and generalization across datasets.

\subsection{3D Object Detection}

For the 3D object detection task, we trained 6 models, each using a different dataset while keeping the hyperparameters consistent across all models. Table~\ref{3d-detection-datasets} presents the volumes of the training datasets used for each of the 6 models.

For training the 3D object detection model SECOND we use the same strategy in 2D object detection and initialize the model weights randomly. We use the AdamW optimizer with initial learning rate 0.001 and a learning rate schedular "StepLR" in pytorch with step size 2 and gamma 0.85. We also use weight decay with 0.0001 and default data augmentation from the OpenPCDet codebase. The batch size is 16 per GPU (64 in total) and we train all of 6 models for 50 epochs with 4 Nvidia RTX 6000 GPUs. All models are converged in 50 epochs. We again observe that 50 epochs are sufficient for all models to converge with respect to the validation loss and select the model checkpoint corresponding to the best validation loss.

\begin{table*}[ht]
\centering
\begin{adjustbox}{valign=c} % This ensures table fits better when necessary
\resizebox{\textwidth}{!}{%
\begin{tabular}{|l|l|c|c|c|c|}
\hline
\textbf{Model Name} & \textbf{Dataset} & \textbf{BIT Frames} & \textbf{KITTI Frames} & \textbf{A2D2 Frames} & \textbf{Total Frames} \\
\hline
3D Model 1 & BIT-TS & 4K & 0 & 0 & 4K \\
3D Model 2 & KITTI & 0 & 4K & 0 & 4K \\
3D Model 3 & BIT-TS \& KITTI MIX & 2K & 4K & 0 & 6K \\
3D Model 3+ & BIT-TS \& KITTI MIX+ & 4K & 4K & 0 & 8K \\
3D Model 4 & A2D2 & 0 & 0 & 4K & 4K \\
3D Model 4+ & BIT-TS \& A2D2 MIX+ & 4K & 0 & 4K & 8K \\
\hline
\end{tabular}
}
\end{adjustbox}
\caption{3D Object Detection - SECOND: Training datasets for the 6 models (Note: "K" denotes thousands).}
\label{3d-detection-datasets}
\end{table*}

\begin{figure*}[h!]
    \centering
    \begin{subfigure}[b]{0.45\textwidth}
        \centering
        \begin{tikzpicture}
            \begin{axis}[
                width=\textwidth,
                height=6cm,
                ybar=2pt,
                xtick=data,
                symbolic x coords={BIT-TS, KITTI, BIT-TS \& KITTI MIX, BIT-TS \& KITTI MIX+, A2D2, BIT-TS \& A2D2 MIX+},
                ylabel={mAP@50},
                xlabel={Training Dataset},
                bar width=12pt,
                enlarge x limits=0.15,
                grid=major,
                yticklabel style={/pgf/number format/.cd, fixed, precision=0},
                nodes near coords,
                nodes near coords style={font=\footnotesize},
                x tick label style={rotate=45,anchor=north east,font=\footnotesize},
                title={3D Det. Performance on KITTI},
                ymin=0,
                ymax=100
            ]
            \addplot coordinates {(BIT-TS,19.9) (KITTI,48.0) (BIT-TS \& KITTI MIX,45.6) (BIT-TS \& KITTI MIX+,45.2) (A2D2,21.7) (BIT-TS \& A2D2 MIX+,25.9)};
            \end{axis}
        \end{tikzpicture}
        \caption{KITTI Point Cloud Test Set Results}
        \label{fig:3d-kitti}
    \end{subfigure}
    \hfill
    \begin{subfigure}[b]{0.45\textwidth}
        \centering
        \begin{tikzpicture}
            \begin{axis}[
                width=\textwidth,
                height=6cm,
                ybar=2pt,
                xtick=data,
                symbolic x coords={BIT-TS, KITTI, BIT-TS \& KITTI MIX, BIT-TS \& KITTI MIX+, A2D2, BIT-TS \& A2D2 MIX+},
                ylabel={mAP@50},
                xlabel={Training Dataset},
                bar width=12pt,
                enlarge x limits=0.15,
                grid=major,
                yticklabel style={/pgf/number format/.cd, fixed, precision=0},
                nodes near coords,
                nodes near coords style={font=\footnotesize},
                x tick label style={rotate=45,anchor=north east,font=\footnotesize},
                title={3D Det. Performance on BDD100K},
                ymin=0,
                ymax=100
            ]
            \addplot [fill=red] coordinates {(BIT-TS,20.5) (KITTI,22.3) (BIT-TS \& KITTI MIX,23.7) (BIT-TS \& KITTI MIX+,26.1) (A2D2,34.4) (BIT-TS \& A2D2 MIX+,27.5)};
            \end{axis}
        \end{tikzpicture}
        \caption{BDD100K Point Cloud Test Set Results}
        \label{fig:3d-a2d2}
    \end{subfigure}
    \caption{Test set 3D object detection mAP@50 values for SECOND - 3D Detection Models}
    \label{fig:3d-detection-results}
\end{figure*}
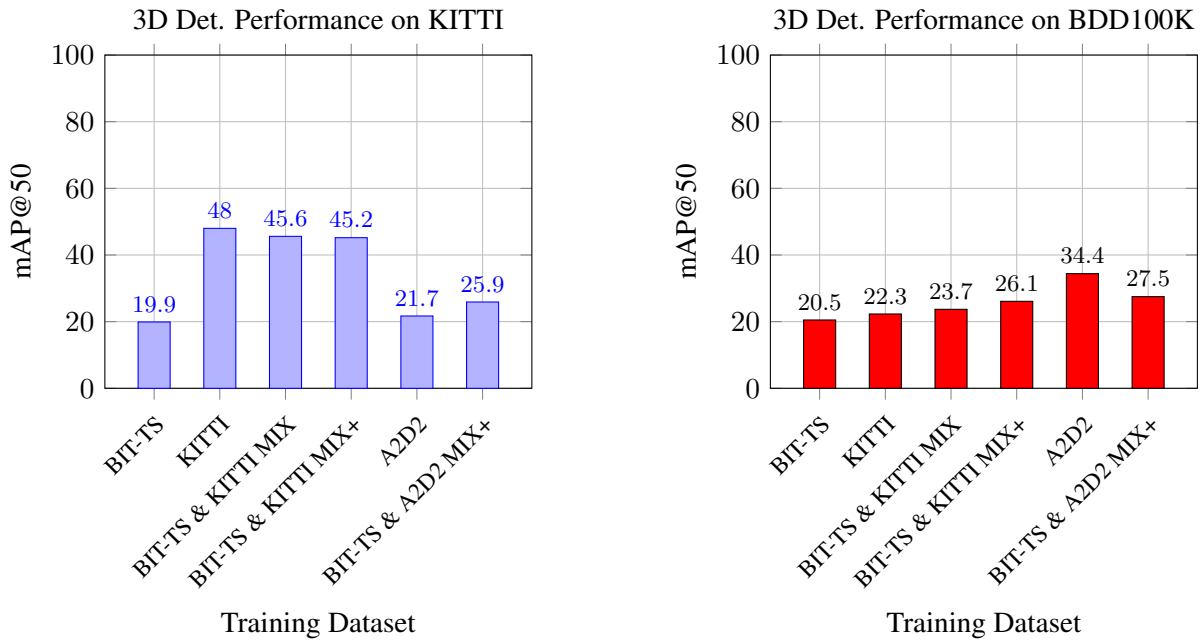

In Figure~\ref{fig:3d-detection-results}, the test results on the KITTI and A2D2 datasets are presented. The test sets were again held out from the training and validation sets.

Unlike the findings in the 2D object detection experiments, the results for 3D object detection on point cloud data reveal that adding synthetic data to real data reduces performance on the same real data test set. This outcome is likely due to the distinct patterns and biases present in the LiDAR point clouds of each dataset. The differences in distributions mean that incorporating synthetic data shifts the model away from the distribution of the real dataset, leading to decreased performance (see Figure~\ref{fig:3d-detection-results}).

However, the model trained on the BIT-TS \& A2D2 MIX+ dataset performs better than the one trained solely on the A2D2 dataset on the KITTI test set. This finding indicates that adding synthetic data to real data still enhances the model's robustness and generalization capabilities. This observation aligns with the conclusions drawn from the 2D object detection experiments, highlighting the potential benefits of leveraging synthetic data for improving generalization across different datasets.

\section{Challenges and Limitations}

In our experiments on both 2D and 3D object detection tasks, we observed that using a mixture of synthetic and real data enhances model robustness. However, the datasets available for our experiments were limited in size, as outlined in Table~\ref{datasets-and-volumes}. In contrast, current autonomous driving systems are typically trained on much larger datasets. Conducting similar experiments at such scales remains an open question and presents a significant area for future research.

Additionally, the scope of this study was restricted to pedestrians and cars. Advanced autonomous driving systems must recognize a far greater variety of objects to be fully effective. Expanding the range of object categories in future experiments could lead to different results and insights.

Finally, while we employed well-established open-source algorithms, the rapid development of state-of-the-art (SoTA) methods introduces another potential limitation. The outcomes of these experiments may be influenced by the choice of model, and evaluating newer algorithms could provide further understanding. Addressing these aspects in future studies will be critical to advancing the field.

\section{Conclusion}

This study investigates the impact of synthetic data on object detection performance for both 2D and 3D tasks. By training models on real, synthetic, and mixed datasets, we evaluated their ability to generalize across diverse scenarios.

The results demonstrate that synthetic data, particularly when combined with real-world data, significantly enhances model performance and generalization. For 2D object detection, models trained with a mix of real and synthetic datasets consistently outperformed those trained exclusively on either type of data. The same trend was observed in 3D object detection, where incorporating synthetic data improved performance across test datasets. Notably, the BIT-TS synthetic dataset proved to be a versatile source of additional training data, adapting well to diverse scenarios and bridging gaps in real-world datasets.

These findings underline the potential of synthetic data as a complementary resource for real-world datasets in object detection tasks. Synthetic data not only augments the diversity of training samples but also mitigates the challenges posed by data scarcity in certain domains.

\section*{Acknowledgements}
We extend our heartfelt gratitude to Yusuf Güleray and Ivan Maksimenko for their insightful discussions and valuable feedback.

\bibliography{testbench}

\begin{thebibliography}{10}
\expandafter\ifx\csname natexlab\endcsname\relax\def\natexlab#1{#1}\fi

\bibitem[{Geiger et~al.(2013)Geiger, Lenz, Stiller, and Urtasun}]{geiger2013vision}
Andreas Geiger, Philip Lenz, Christoph Stiller, and Raquel Urtasun. 2013.
\newblock Vision meets robotics: The kitti dataset.
\newblock \emph{The International Journal of Robotics Research}, 32(11):1231--1237.

\bibitem[{Geyer et~al.(2020)Geyer, Kassahun, Mahmudi, Ricou, Durgesh, Chung, Hauswald, Pham, M{\"u}hlegg, Dorn et~al.}]{geyer2020a2d2}
Jakob Geyer, Yohannes Kassahun, Mentar Mahmudi, Xavier Ricou, Rupesh Durgesh, Andrew~S Chung, Lorenz Hauswald, Viet~Hoang Pham, Maximilian M{\"u}hlegg, Sebastian Dorn, et~al. 2020.
\newblock A2d2: Audi autonomous driving dataset.
\newblock \emph{arXiv preprint arXiv:2004.06320}.

\bibitem[{Grigorescu et~al.(2020)Grigorescu, Trasnea, Cocias, and Macesanu}]{grigorescu2020survey}
Sorin Grigorescu, Bogdan Trasnea, Tiberiu Cocias, and Gigel Macesanu. 2020.
\newblock A survey of deep learning techniques for autonomous driving.
\newblock \emph{Journal of field robotics}, 37(3):362--386.

\bibitem[{Redmon(2018)}]{redmon2018yolov3}
Joseph Redmon. 2018.
\newblock Yolov3: An incremental improvement.
\newblock \emph{arXiv preprint arXiv:1804.02767}.

\bibitem[{Song et~al.(2023)Song, He, Li, Ma, Ming, Mao, Pei, Peng, Hu, Yao et~al.}]{song2023synthetic}
Zhihang Song, Zimin He, Xingyu Li, Qiming Ma, Ruibo Ming, Zhiqi Mao, Huaxin Pei, Lihui Peng, Jianming Hu, Danya Yao, et~al. 2023.
\newblock Synthetic datasets for autonomous driving: A survey.
\newblock \emph{IEEE Transactions on Intelligent Vehicles}.

\bibitem[{Talwar et~al.(2020)Talwar, Guruswamy, Ravipati, and Eirinaki}]{talwar2020evaluating}
Deepak Talwar, Sachin Guruswamy, Naveen Ravipati, and Magdalini Eirinaki. 2020.
\newblock Evaluating validity of synthetic data in perception tasks for autonomous vehicles.
\newblock In \emph{2020 IEEE International Conference On Artificial Intelligence Testing (AITest)}, pages 73--80. IEEE.

\bibitem[{Team(2020)}]{openpcdet2020}
OpenPCDet~Development Team. 2020.
\newblock Openpcdet: An open-source toolbox for 3d object detection from point clouds.
\newblock \url{https://github.com/open-mmlab/OpenPCDet}.

\bibitem[{Wang et~al.(2023)Wang, Bochkovskiy, and Liao}]{wang2023yolov7}
Chien-Yao Wang, Alexey Bochkovskiy, and Hong-Yuan~Mark Liao. 2023.
\newblock Yolov7: Trainable bag-of-freebies sets new state-of-the-art for real-time object detectors.
\newblock In \emph{Proceedings of the IEEE/CVF conference on computer vision and pattern recognition}, pages 7464--7475.

\bibitem[{Yan et~al.(2018)Yan, Mao, and Li}]{yan2018second}
Yan Yan, Yuxing Mao, and Bo~Li. 2018.
\newblock Second: Sparsely embedded convolutional detection.
\newblock \emph{Sensors}, 18(10):3337.

\bibitem[{Yu et~al.(2020)Yu, Chen, Wang, Xian, Chen, Liu, Madhavan, and Darrell}]{yu2020bdd100k}
Fisher Yu, Haofeng Chen, Xin Wang, Wenqi Xian, Yingying Chen, Fangchen Liu, Vashisht Madhavan, and Trevor Darrell. 2020.
\newblock Bdd100k: A diverse driving dataset for heterogeneous multitask learning.
\newblock In \emph{Proceedings of the IEEE/CVF conference on computer vision and pattern recognition}, pages 2636--2645.

\end{thebibliography}
\bibliographystyle{acl_natbib}

\end{document}